\documentclass[10pt,twocolumn,letterpaper]{article}

\usepackage{cvpr}              

\usepackage{graphicx}
\usepackage{amsmath}
\usepackage{amssymb}
\usepackage{booktabs}
\usepackage[pagebackref,breaklinks,colorlinks]{hyperref}
\usepackage[table]{xcolor}
\usepackage{multirow}
\usepackage{tabularx}
\usepackage[font=small,labelfont=bf,tableposition=top]{caption}
\DeclareCaptionLabelFormat{andtable}{#1~#2  \&  \tablename~\thetable}
\usepackage{calc}

\usepackage{blindtext}

\newcommand{\model}{SUNDIAL}
\definecolor{mygreen}{HTML}{E2F4E4}
\definecolor{tabfirst}{rgb}{0.87, 1, 0.87} 
\definecolor{tabsecond}{rgb}{1, 1, 1} 
\definecolor{tabthird}{rgb}{1, 1, 1} 

\newcommand{\prio}{\mathbf{o}}
\newcommand{\prid}{\mathbf{d}}
\newcommand{\prir}{\mathbf{r}}
\newcommand{\seco}{\mathbf{o}_{\text{s}}}
\newcommand{\secd}{\mathbf{d}_{\text{s}}}
\newcommand{\secr}{\mathbf{s}}
\newcommand{\pdi}{P_{\text{di}}}

\usepackage[capitalize]{cleveref}
\crefname{section}{Sec.}{Secs.}
\Crefname{section}{Section}{Sections}
\Crefname{table}{Table}{Tables}
\crefname{table}{Tab.}{Tabs.}

\def\cvprPaperID{13496} 
\def\confName{CVPR}
\def\confYear{2024}

\begin{document}

\title{SUNDIAL: 3D \underline{S}atellite \underline{Un}derstanding through \\ \underline{Di}rect, \underline{A}mbient, and Complex \underline{L}ighting Decomposition}
\newcommand{\superscript}[1]{\ensuremath{^{\textrm{#1}}}}
\author{
    Nikhil Behari\superscript{1},
    Akshat Dave\superscript{2},
    Kushagra Tiwary\superscript{2},
    William Yang\superscript{2},
    Ramesh Raskar\superscript{2}\\
    \superscript{1}Harvard University, \superscript{2}Massachusetts Institute of Technology\\
    {\tt\small nikhilbehari@g.harvard.edu, \{ad74,ktiwary,wyyang,raskar\}@mit.edu}
}

\twocolumn[{%
\renewcommand\twocolumn[1][]{#1}%
\maketitle
\begin{center}
    \centering
    \captionsetup{type=figure}
    \label{fig:teaser}
    \includegraphics[width=0.95\textwidth]{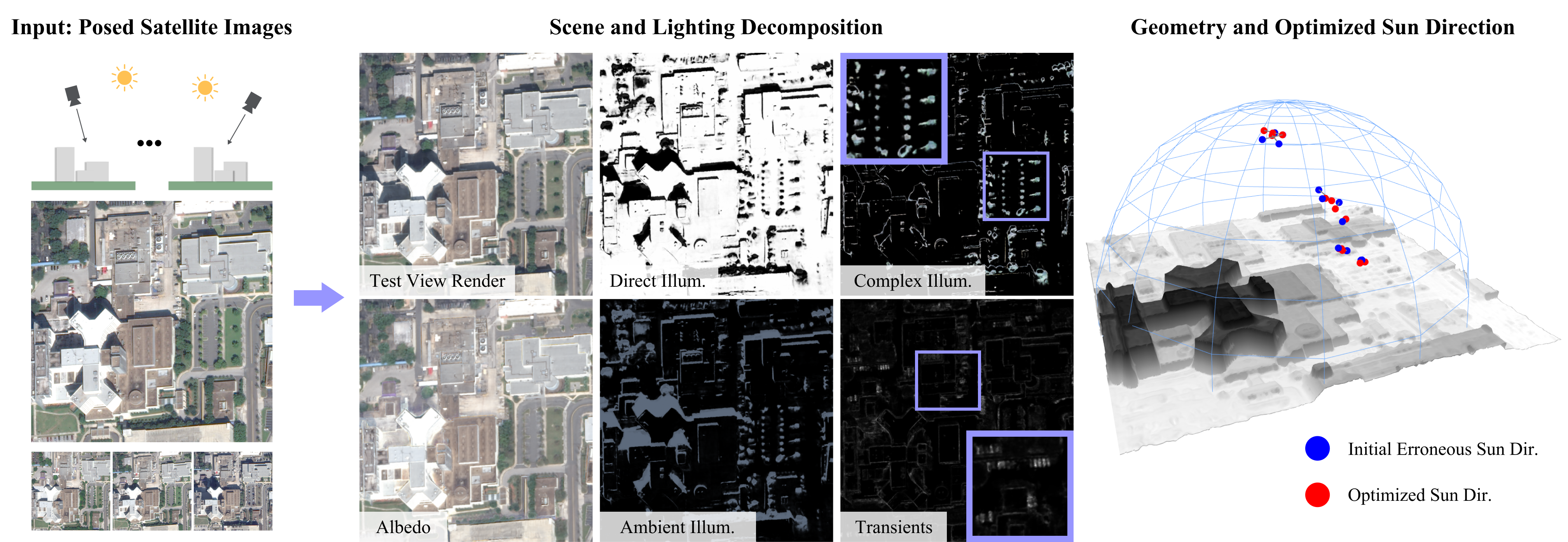}
    \captionof{figure}{\textbf{\model{} scene and lighting disentanglement}. \model{} resolves ambiguous direct, ambient, and complex illumination effects and estimates accurate surface albedo and transient features in satellite imagery, resulting in high-quality renders under novel viewpoints and sun positions. By adapting priors from remote sensing literature, \model{} adheres to a more physically-accurate scene rendering, and enables joint estimation of scene geometry and sun direction. 
    }
\end{center}%
}]
\maketitle

\begin{abstract}
   3D modeling from satellite imagery is essential in areas of environmental science, urban planning, agriculture, and disaster response. However, traditional 3D modeling techniques face unique challenges in the remote sensing context, including limited multi-view baselines over extensive regions, varying direct, ambient, and complex illumination conditions, and time-varying scene changes across captures. In this work, we introduce \model{}, a comprehensive approach to 3D reconstruction of satellite imagery using neural radiance fields. We jointly learn satellite scene geometry, illumination components, and sun direction in this single-model approach, and propose a secondary shadow ray casting technique to 1) improve scene geometry using oblique sun angles to render shadows, 2) enable physically-based disentanglement of scene albedo and illumination, and 3) determine the components of illumination from direct, ambient (sky), and complex sources. To achieve this, we incorporate lighting cues and geometric priors from remote sensing literature in a neural rendering approach, modeling physical properties of satellite scenes such as shadows, scattered sky illumination, and complex illumination and shading of vegetation and water. We evaluate the performance of \model{} against existing NeRF-based techniques for satellite scene modeling and demonstrate improved scene and lighting disentanglement, novel view and lighting rendering, and geometry and sun direction estimation on challenging scenes with small baselines, sparse inputs, and variable illumination. 
\end{abstract}



\section{Introduction}
\label{sec:intro}
Every day, 1.3 million images are taken from satellites orbiting the Earth, with an estimated 1,200 images captured for every point on the Earth’s landmass \cite{Mohney_2020}. This imagery has wide-scale applications across fields of environmental science, urban development, agriculture, and disaster response. Extracting 3D information from this remote sensing imagery, in particular, remains an open problem with significant impact in each field. For example, 3D scene understanding can help provide greater spatial awareness and context while monitoring global phenomena such as deforestation, climate change, and urban expansion. 

3D satellite scene reconstruction, however, presents multiple distinct challenges. First, satellite captures typically have limited multi-view baselines for vast capture regions; multi-view cues may only be visible in off-nadir captures from low-altitude, high-resolution satellite imagery \cite{longbotham2011spatial}. Second, multi-view satellite captures can be taken over time, and therefore contain transient, or changing, features such as shifting cars or ongoing construction. Third, illumination in satellite imagery includes components of direct, atmospheric, reflected, and scattered irradiance \cite{adeline2013shadow}; each component can vary greatly across captures, causing inconsistencies across image inputs. Fourth, accurately estimating camera and lighting poses can be challenging, especially given the small baselines and large scene distances; small errors in these poses can propagate through 3D modeling resulting in large geometry estimation errors \cite{mari2019bundle}. Despite these challenges, properly disentangling 3D scene geometry and illumination components is essential to minimize ambiguity and loss of spatial, lighting, and temporal information. \cite{su2016shadow,wang2019detection,liu2012object}.

In order to overcome the unique constraints of 3D satellite imagery, shadow-based height estimation has been proposed to overcome insufficient multi-view samples, harnessing oblique sun angles and shadow rendering to uncover hidden scene geometry \cite{irvin1989methods,liasis2016satellite}. Recent techniques have applied neural radiance fields (NeRFs) \cite{mildenhall2021nerf} to model and separate these shadows in satellite imagery. S-NeRF \cite{derksen2021shadow} first proposed estimating a sun-dependent shading scalar across images; SatNeRF \cite{mari2022sat} extended this work, modeling transient features in addition to shading. EO-NeRF \cite{mari2022sat} then proposed a geometric NeRF-based shadow rendering approach, casting secondary sun-directed rays to determine regions with primary occluders (i.e. in shadow). However, these existing methods largely overlook the unique aspects of illumination specific to satellite-captured scenes. In particular, multi-bounce irradiance and reflections can create complex shading and soft shadows that cannot be modeled with a secondary ray technique. As a result, these approaches struggle to accurately decompose illumination and scene geometry (Figure \ref{fig:can_sat_comp}), causing ambiguities in separation. Overcoming these ambiguities is critical; accurate shadow separation in satellite imagery can help identify occluded objects \cite{su2016shadow}, enable more accurate land cover classification \cite{liu2012object}, and uncover shadow-related surface temperature cooling effects \cite{yu2019study}. 


In this work, we present \model{}, a NeRF-based framework that reduces ambiguity in satellite scene decomposition by estimating three key illumination components: direct illumination, ambient illumination, and complex illumination. We propose a geometric shadow ray casting approach, using the transmittance of these rays to render physically-accurate, disentangled shadows. However, in addition to estimating direct and indirect illumination component, we use these secondary rays to predict an additional complex illumination component that captures the challenging, multi-bounce light transport of satellite scenes that may be computationally infeasible to explicitly model. By estimating this additional scene lighting component, we are also able to reduce ambiguity in shadow rendering, thereby improving estimated scene geometry and lighting decomposition in satellite-captured scenes. 

We propose several techniques to achieve this scene decomposition. First, we use secondary shadow ray transmission to estimate the components of illumination at a scene point, and apply loss functions inspired by prior research in remote sensing to accurately predict ambient and complex illumination. Second, we enforce more confident shadow predictions, thereby refining both scene geometry and shadow rendering jointly, and apply geometric regularization to improve scene modeling. Third, given the critical importance of direct illumination and shadow rendering in our approach, we jointly estimate sun direction with scene geometry during training, enabling refined sun direction prediction and mitigating propagated error from incorrect initial poses.

In summary, our contributions are the following: 
\begin{itemize}
    \item We disentangle and estimate direct, ambient, and complex illumination in satellite imagery, enabling more accurate, physically-based 3D scene modeling 
    \item We jointly estimate sun directions and scene geometry, and propose remote sensing-based illumination and geometric priors to improve scene decomposition
    \item We enable more accurate separation of satellite scene albedo, shadow, and complex shading through our decomposition technique, improving novel view and lighting synthesis for satellite imagery
\end{itemize}


\section{Related Work}
\label{sec:relatedwork}

\subsection{Neural Radiance Fields}
Neural Radiance Fields (NeRFs) have gained significant traction in the field of multi-view 3D reconstruction. By modelling the scene as coordinate-based neural networks, Mildenhall et al. \cite{mildenhall2021nerf} demonstrated that NeRFs can synthesize novel views of complex scenes with unprecedented detail and photo-realism. 
Subsequent works have scaled NeRFs from smartphone captures to in-the-wild internet images \cite{martin2021nerf}, large unbounded scenes \cite{barron2022mip}, drone videos \cite{barron2023zip}, city-scale regions \cite{tancik2022block,xiangli2022bungeenerf}, and satellite imagery \cite{mari2022sat,xie2023snerf}. NeRFs also show significant promise in 3D computer vision, with applications ranging from simultaneous localization and mapping \cite{zhu2022nice}, robotics control \cite{li20223d}, manipulation \cite{simeonov2022neural} and navigation \cite{adamkiewicz2022vision}, semantic segmentation \cite{vora2021nesf} to scene understanding \cite{zhi2021place}.

\noindent\textbf{Disentangling appearance in NeRFs}
Conventional NeRFs model the total radiance emitted by the scene that comprises of the lighting, geometry and material properties of the scene. Several works have explored disentangling the emitted radiance into its constituents by leveraging reflectance priors \cite{boss2021nerd,boss2021neural,zhang2021physg,zhang2021nerfactor}, specular reflections \cite{verbin2022ref,tiwary2023orca,liu2023nero}, shading \cite{bi2020neural,zhang2022iron} and shadow cues \cite{yang2022s,tiwary2022towards}. In this work, we focus on shadow-based disentanglement given the prominence of hard shadows in satellite imagery. We show how transmittance along the shadow ray can allow us to decompose not just the direct and shadow components, but also the components corresponding to ambient environment illumination and complex lighting effects.

\subsection{Remote Sensing}
Remote sensing research has advanced significantly in recent decades, yet there remain key challenges in building generalized, illumination-aware 3D reconstructions from satellite imagery. Particularly in urban areas, large shadows from buildings can cause significant ambiguity and classification errors \cite{dare2005shadow,weng2012remote}. Both thresholding and geometric modeling techniques have been applied to this problem of shadow removal \cite{rs13040699,liu2012object}. Multi-view stereo (MVS) is another standard approach to 3D scene modeling, utilizing stereo pairs across images \cite{gomez2022experimental,berra2020advances}. However, MVS can produce errors due to variability in sun direction, atmospheric conditions, and seasonal features across images \cite{Cheng_2020,mostafa2017review,su2016shadow}, requiring manual filtering of stereo pairs \cite{gomez2022experimental,facciolo2017automatic}. 

\noindent\textbf{3D Satellite Modeling with NeRF} Neural radiance fields (NeRFs) have recently been applied to overcome these challenges. S-NeRF \cite{derksen2021shadow} introduced a technique for separating shading from albedo; Sat-NeRF \cite{mari2022sat} extended this work, highlighting the importance of bundle adjustment for improving reconstruction. EO-NeRF \cite{mari2023multi} further implemented a geometric shadow rendering approach, casting secondary rays in input sun directions to refine geometry. However, these methods still result in scene disentangling ambiguities: S-NeRF and SatNeRF estimate scene shading irrespective of geometry, resulting in inaccuracies in separation; EO-NeRF uses a geometry-based shadow rendering but still captures scene shading in non-shadowed regions, indicating that separating direct and indirect lighting alone is insufficient for remote sensing reconstruction. By modeling complex illumination effects, sun position estimation, and geometry priors, we achieve more accurate, physically-based scene and lighting disentanglement compared to prior NeRF-based techniques for satellite imagery (Fig. \ref{fig:can_sat_comp}). 

\begin{figure*}[t]
    \centering
    \includegraphics[width=0.9\linewidth]{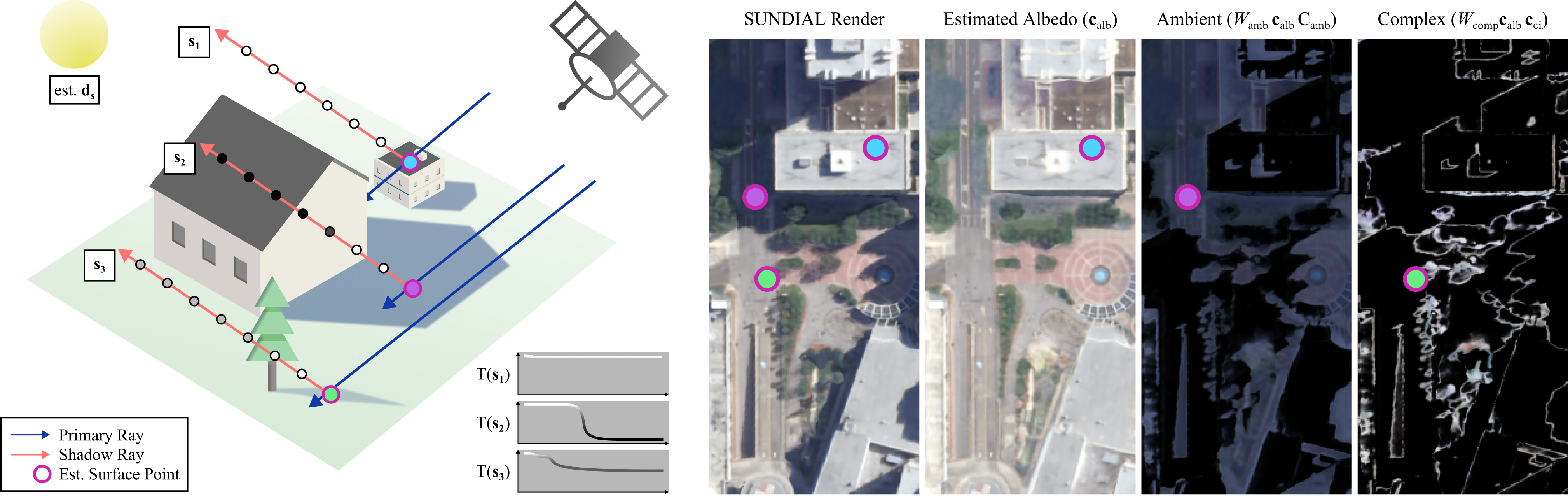}
        \caption{\textbf{Accumulated transmittance from secondary shadow rays as illumination cues}. Final transmittance values of secondary shadow rays are used to determine the proportion of direct, ambient, and complex illumination values. Direct light is the primary component of rays with direct sun visibility (e.g. ray $\secr{}_1$); ambient light is the primary component of rays distinctly in shadow (e.g. ray $\secr{}_2$); complex lighting is the primary component of rays with ambiguous or challenging geometry occluders (e.g. ray $\secr{}_3$). }
    \label{fig:secondaryraycasting}
\end{figure*}

\section{Formulation}
\label{sec:formulation}

\subsection{Small Baselines and Hidden Regions}
There are several challenges unique to 3D satellite image reconstruction; most notably, images are captured from a top-down perspective with extremely small multi-view baselines, making it difficult to capture vertical surface geometries such as building facades \cite{tiwary2022towards,zhao2023review}. To account for these regions with insufficient primary ray samples, we cast secondary shadow rays drawn from scene surface points to the direct lighting source (i.e. the sun). While vertical surfaces may be hidden from primary satellite capture angles, oblique lighting angles from the sun can help exploit and model this hidden geometry.

We use these secondary shadow rays for three key purposes. First, these rays are used to refine scene geometry; we cast secondary rays from estimated origin points on the visible scene surface in the direction of the sun (Figure \ref{fig:secondaryraycasting}), rendering shadows from these rays to improve estimated scene geometry. Second, we render geometry and sun-dependent shadows using these secondary rays, helping to separate scene albedo from shadows and ambient lighting in the final rendering approach. Third, we use these secondary shadow rays to determine the primary illumination components for a given pixel. Specifically, we propose using the final transmittance of these secondary rays to determine whether a region is either primarily in direct sun lighting, primarily in shadow, or a more complex combination of multi-bounce light that cannot be directly modeled with a single-bounce secondary ray casting. 

\subsection{Disentangling Illumination}
Previous studies identify four primary radiative components in satellite imagery: 1) direct solar irradiance, 2) scattered atmospheric irradiance, 3) irradiance from reflection between objects, and 4) coupled irradiance from multiple scattering events \cite{adeline2013shadow}. To better disentangle these lighting components, we use secondary \model{} shadow rays to estimate the weights of three illumination components for a given ray $\prir{}$: 

\begin{enumerate}
    \item $W_{\text{di}}(\prir{})$: Direct illumination component weight
    \item $W_{\text{amb}}(\prir{})$: Ambient illumination component weight
    \item $W_\text{comp}(\prir{})$: Complex illumination component weight
\end{enumerate}

In general, $W_{\text{di}}$ is highest for regions with direct sun visibility, $W_{\text{amb}}$ is highest in shadowed areas with strong atmospheric irradiance, and $W_\text{comp}$ is highest in complex-geometry regions such as trees and water, where interreflections and multiple scattering events create intricate shading \cite{shepherd2003correcting,guo2010removing,bishop2011topographic}. Secondary shadow rays are used to determine weightings for each illumination component, which we apply to three predicted rendering colors. 


\noindent\textbf{Direct Illumination}
Direct illumination is the primary irradiance component for regions with direct sun visibility. In these regions, the primary component of our final color rendering is predicted as position-dependent $\mathbf{c}_{\text{alb}}(\mathbf{x})$, the albedo color. We compute this direct lighting component in the final rendering as: 

\begin{equation}
    W_{\text{di}}(\prir{}) \cdot \mathbf{c}_{\text{alb}}
\end{equation}

Which is a weighted quantity representing the direct illumination component of the final ray color. For pixels in complete direct illumination (i.e. not in shadow), $W_{\text{di}}(\prir{}) = 1$, and the predicted albedo color is therefore the primary color component of the final rendering. 

\noindent\textbf{Ambient Illumination} 
For regions completely in shadow, direct solar radiation is blocked, so scattered atmospheric irradiance contributes most to total irradiance \cite{zhu2018automatic,adeline2013shadow}. To model this atmospheric irradiance, we predict an ambient color $C_{\text{amb}}(\secd{})$. This color is a function of the estimated sun position $\secd{}$ for simplicity; in reality, atmospheric scattering may depend on a variety of additional scatting features, such as air, smoke, and cloud particles \cite{polidorio2003automatic}.  This value $C_{\text{amb}}(\secd{})$ is a single RGB color value (typically a dark blue hue) multiplied by the base albedo to render regions in shadow where atmospheric irradiance is strongest. We therefore apply this ambient light as a weighted component (strongest for regions in shadow): 

\begin{equation}
    W_{\text{amb}}(\prir{}) \cdot \mathbf{c}_{\text{alb}} \otimes C_{\text{amb}}
\end{equation}

where $\otimes$ denotes elementwise multiplication. 

\noindent\textbf{Complex Illumination}
Captured irradiance in satellite imagery may also be explained by more complex multi-bounce illumination features such as interreflections and scattered light. Thus, we additionally predict a ``complex illumination" feature $\mathbf{c}_{\text{ci}}(\mathbf{x}, \secd{})$ as a function of position and sun direction. This complex illumination feature $\mathbf{c}_{\text{ci}}$ is applied in the final rendering equation with weighted quantity $W_{\text{comp}}(\prir{})$, which is highest when occluder geometry is ambiguous (i.e. in trees, bushes, and water). We apply this spatially and sun position-dependent complex illumination component in the final rendering as: 

\begin{equation}
W_{\text{comp}}(\prir{}) \cdot \mathbf{c}_{\text{alb}} \cdot \mathbf{c}_{\text{ci}} 
\end{equation}

\subsection{Final Rendering Equation}

We combine the direct, indirect, and complex illumination components to form the final rendering equation: 

\vspace{-1em} 

\begin{equation}
\mathbf{C}(\prir{}) = W_{\text{di}} \mathbf{c}_{\text{alb}} + W_{\text{amb}}\mathbf{c}_{\text{alb}} \otimes C_{\text{amb}} + W_{\text{comp}} \mathbf{c}_{\text{alb}} \mathbf{c}_{\text{ci}} 
\end{equation}

In this final rendering equation, the first term is strongest for regions with well-defined geometry and that are directly visible to the sun. The second term is strongest for regions with well-defined geometry that are entirely in shadow (occluded from direct sun visibility). The third term is strongest for regions requiring multi-bounce light shading, for example in regions with trees or water; in this case, the complex illumination color can provide nuanced shading, potentially capturing interreflection and multiple scattering components.


\subsection{Refining Poses}
Accurate camera poses are critical for 3D satellite scene reconstruction, especially given the large distance from camera to scene and small multi-view baselines. Prior work has demonstrated the importance of bundle adjustments for the 3D reconstruction task in satellite imagery \cite{mari2022sat,grodecki2003block,ozcanli2014automatic}. We rely on direct sun illumination and shadows as a secondary geometry cue; because camera pose and relative sun direction are closely related, we propose a joint geometry and sun direction estimation technique in \model{}. Our geometric shadow rendering approach relies on accurate shadows to 1) refine geometry, 2) separate and render shadows, and 3) determine illumination components. Thus, we find that jointly estimating sun direction can help avoid propagating error from imprecise estimated sun direction inputs. 


\section{Our Approach}
\label{sec:approach}

\begin{figure}
    \centering
    \includegraphics[width=0.97\columnwidth]{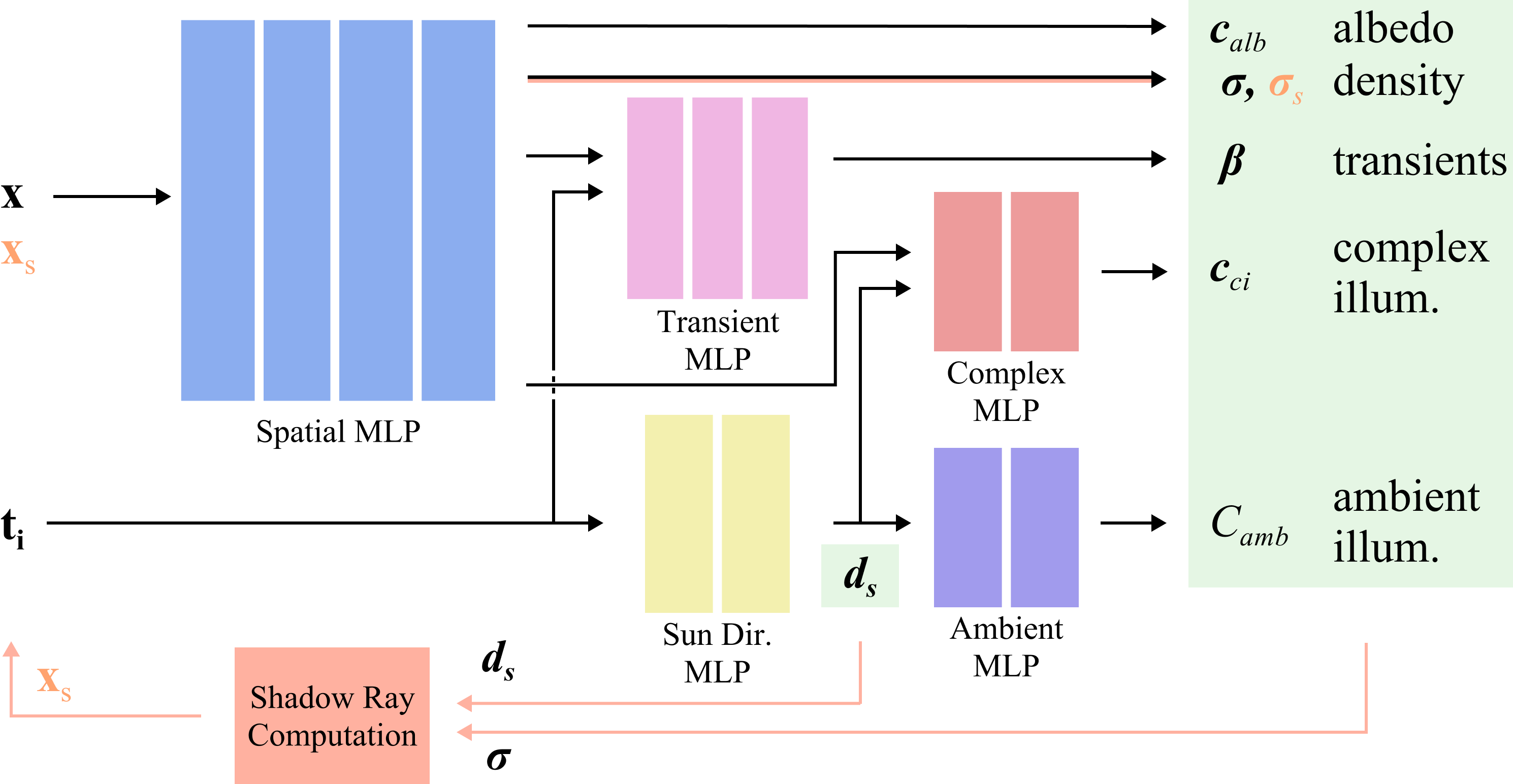}
        \caption{\textbf{Overview of our proposed architecture}. Outputs of \model{} are highlighted in \colorbox{mygreen}{green}. Primary ray densities $\sigma$ are used with estimated sun directions to query secondary shadow rays with density $\sigma_{\text{s}}$; final transmittance values for these secondary ray are used to compute direct, ambient, and complex illumination weights.}
    \label{fig:overview}
\end{figure}



\subsection{NeRF Preliminaries}
\label{subsec:prelim}
NeRF \cite{mildenhall2021nerf} is a volumetric representation, with inputs spatial position $\mathbf{x}$ and viewing direction $\prid{}$, and outputs volume density $\sigma$ and color $\mathbf{c}$: $F: (\mathbf{x}, \prid{}) \rightarrow (\sigma, \mathbf{c})$. Volumetric rendering is achieved by uniformly sampling along rays $\prir{}(t) = \prio{} + t\prid{}$, enabling color $C(\prir{})$ and depth $d(\prir{})$ estimation of a ray as: 

\begin{equation}
C(\prir{}) = \sum_{i} w_i \mathbf{c}_i \quad d(\prir{}) = \sum_i w_i t_i
\end{equation}

\vspace{-1.5em} 

\begin{equation}
T_i = \text{exp}(-\sum_{j=1}^{i=1} \sigma_j \delta_j) \quad w_i = T_i (1-\text{exp}(-\sigma_i\delta_i))  
\end{equation}

Where $T_i$ is the accumulated transmittance by step size $\delta_i$, the distance between sampled points.

\subsection{Primary Ray Components}
\label{subsec:prim_ray}

\noindent \textbf{Albedo} We model scene albedo $\mathbf{c}_{\text{alb}}$ using a coordinate-based MLP. We exclude view dependence in the spatial MLP inputs, assuming Lambertian surface properties. Although non-Lambertian modeling is critical for topography-influenced lighting variation in certain terrains \cite{santosa2016evaluation}, assuming Lambertian surfaces is often suitable for urban aerial imagery \cite{bonafoni2020albedo,hofierka2021estimating}, the primary target of our evaluation. 

\noindent \textbf{Ambient Color} Ambient color $C_{\text{amb}}$, the primary color component for regions in shadow, is predicted using a 2-layer MLP with input the estimated sun directions $\secd{}$. This single RGB color prediction is therefore uniform for a given input image. 

\noindent \textbf{Complex Color} We further model the complex color $\mathbf{c}_{\text{ci}}$, the primary color component for regions with challenging-geometry occluders, using a 2-layer MLP with input spatial features and estimated sun directions $\secd{}$. This modeling technique aims to capture intricate shading, multi-bounce reflections, and ambient occlusions that are difficult to capture with direct and indirect light modeling alone. Instead, shading is applied using coordinate and sun position-based complex color, weighted with estimated quantity $W_{\text{comp}}$. 

\subsection{Sun Direction, Shadows, and Illumination}
\label{subsec:sundir_shads}

For each primary ray $\prir{}$ we cast secondary shadow rays $\secr{}$. These secondary shadow rays are used to determine the proportion of direct, ambient, and complex illumination values. We visualize this shadow ray casting procedure in Figure \ref{fig:secondaryraycasting}.  

\noindent\textbf{Surface Estimation for Shadow Rays} The secondary shadow rays have origin $\seco{}$ estimated using primary ray depth predictions. Given estimated depth $d(\prir{})$ of ray $\prir{}$, we compute a surface point $\seco{}$: 

\vspace{-1em} 

\begin{equation}
\seco{} = \mathbf{o} + (d(\prir{}) - \Delta(t))\prid{}
\end{equation}

Padded by $\Delta(r) = \Delta_0 e^{-kr} $, a function of the training iteration $r$ and $\Delta_0$, some initial small padding constant. In early stages of training, the estimated implicit surface may not be well-formed, resulting in partial initial occlusion in secondary shadow rays. This padding on the estimated depth enables more confident shadow rendering (and downstream sun position estimation) in early stages of training, and decays as the estimated surface becomes more refined. 

\noindent\textbf{Sun Direction Estimation} We estimate sun directions $\secd{}$ jointly with scene geometry during training. We first consider original (input) sun directions $\mathbf{d}_{\text{s0}}$ with initial ``up" directions \( \mathbf{u} \).  This formulation allows for initial estimates to be used for faster convergence when sun direction is known. For training settings without known sun directions, these values may be randomly initialized for each image.

Given an input image embedding \( \mathbf{t}_i \), the \model{} MLP predicts \( \mathbf{u}_i = [u_{i1}, u_{i2}, u_{i3}] \), a new ``up" direction for the rotated sun direction reference frame in image $i$. We compute \( \mathbf{V}_i \), the skew-symmetric matrix from \( \mathbf{v} = \mathbf{u} \times \mathbf{u}_i \), and compute a rotation matrix \( \mathbf{R}_i \) from: 

\vspace{-1em} 

\begin{equation}
\mathbf{V}_i = \begin{pmatrix} 0 & -v_3 & v_2 \\ v_3 & 0 & -v_1 \\ -v_2 & v_1 & 0 \end{pmatrix} \quad \mathbf{R}_i = \mathbf{I} + \mathbf{V}_i + \frac{\mathbf{V}_i^2}{1 + c}
\end{equation}

where \( \mathbf{I} \) is the 3x3 identity matrix, \( c \) is the dot product \( \mathbf{u} \cdot \mathbf{u}_i \), and \( \mathbf{R}_i \) is the final rotation matrix for the sun directions in image $i$. We then obtain the estimated sun direction, used as the direction $\secd{}$ of the secondary shadow rays:
\begin{equation}
    \secd{} = \mathbf{R}_i \cdot \mathbf{d}_{\text{s0}}
\end{equation}

\noindent\textbf{Secondary Shadow Rays} With the network-estimated surface points $\mathbf{o}_{s}$ and predicted sun directions $\secd{}$, we then cast secondary rays $\secr{}$ to predict shadowed regions. We determine the probability of the pixel being in shadow by the transmittance at the end of the secondary shadow ray:


In shadowed regions, this secondary ray transmittance will be low, as the ray $\secr{}$ will be occluded from direct sun visibility. Thus, we obtain the probability that the surface of ray $\prir{}$ is in direct sun lighting as: $\pdi{}(\prir{}) = T(\secr{})$.

\noindent\textbf{Estimating Illumination Weights} Using the probability of direct lighting $\pdi{}(\prir{})$, we then compute the weights for the three illumination components. First, we compute $W_{\text{comp}}$: 

\vspace{-1em} 

\begin{equation}
W_{\text{comp}} = 1 - ( \frac{1}{e^{(\kappa - \xi) - \kappa \pdi{}(\prir{})} + 1} + \frac{1}{e^{\kappa \pdi{}(\prir{}) - \xi} + 1} )
\end{equation}


Where $\kappa$ and $\xi$ are hyperparameters that can be tuned according to the desired sensitivity to complex illumination. This quantity is highest when predicted direct lighting probability $\pdi{}(\prir{})$ is near 0.5, where occluder geometry may be ambiguously defined). Then, we obtain $W_{\text{di}}$ and $W_{\text{amb}}$ as: 


\begin{equation}
    W_{\text{di}} = (1 - W_{\text{comp}}) \cdot \pdi{}(\prir{})
\end{equation}

\vspace{-1em}

\begin{equation}
    W_{\text{amb}} = (1 - W_{\text{comp}}) \cdot (1 - \pdi{}(\prir{}))
\end{equation}

Intuitively, complex illumination weight decreases as $\pdi{}(\prir{})$ probabilities become more confident (i.e. closer to 0 or 1), and increases when predictions are ambiguous (when $\pdi{}(\prir{}) = 0.5$). This weighting method has to key advantages. First, this approach enables more confident shadow predictions, reducing the need for \model{} to handle semi-shaded areas with partial volume occlusions that may negatively impact estimated geometry. Second, this approach represents complex illumination in both directly lit and shadowed regions, increasing complex illumination weight as the precision of geometry-based occlusion prediction decreases.

\begin{figure*}[ht]
    \centering
    \includegraphics[width=0.85\linewidth]{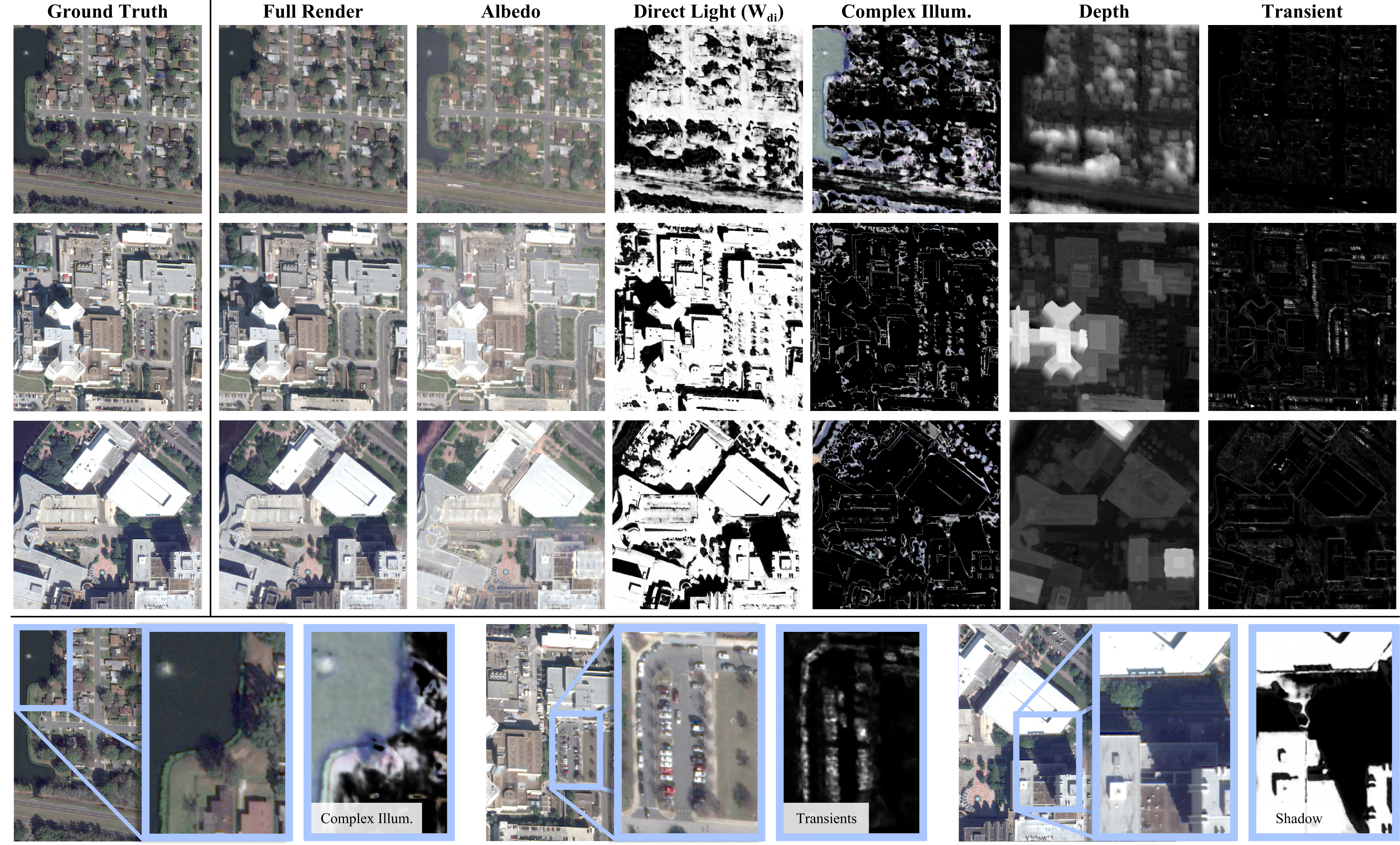}
    \caption{\textbf{\model{} scene disentangling results}. \model{} can disentangle complex illumination (soft shadows, shading) in challenging-geometry regions such as trees, bushes, and water. This complex illumination separation disambiguates albedo, transient, and shadow predictions, refining \model{}-estimated scene decomposition and improving novel viewpoint and sun direction rendering. }
    \label{fig:full_dis}
\end{figure*}

\subsection{Loss Functions}
\label{subsec:loss_funcs}

We incorporate photometric loss, normal orientation loss \cite{verbin2022ref}, and distortion loss \cite{barron2022mip} in our architecture. We additionally introduce three loss functions that help improve rendering specifically for satellite imagery scenes by constraining ambient and complex color predictions. 

\noindent \textbf{Ambient Color Loss}
Prior work in atmospheric modeling indicates that, due to the Rayleigh effect of atmospheric scattering, the ambient lighting component of shadowed regions has a higher saturation of blue and violet colors \cite{zhou2021shadow,polidorio2003automatic}. We therefore apply two physics-based loss functions on the predicted ambient color $C_{amb}$: 

\vspace{-1em} 

\begin{equation}
\mathcal{L}_{\text{white}}(C_{\text{amb}}) = \lambda_{\text{w}} \cdot \max(0, \frac{1}{3}(r + g + b) - \tau )^2
\end{equation}

\vspace{-2em} 

\begin{equation}
\mathcal{L}_{\text{blue}}(C_{\text{amb}}) = \lambda_{\text{b}} \cdot \max(0, \psi \cdot \frac{r + g}{2} - b )^2
\end{equation}

Where $\tau$ is a ``brightness" threshold, $\psi$ is a ``blue dominance" factor, and $\lambda_{\text{w}}$ and $\lambda_{\text{b}}$ are scaling terms. These terms ensure that predicted ambient light color $C_{amb}$ is sufficiently dark and blue in color, adhering to establish principles of atmospheric modeling. We find this prior is especially critical in cases where input images have shadows in similar positions across captures, making it difficult to separate albedo and shadows. The proposed ambient color loss ensures that rendered shadows are sufficiently dark blue (from ambient light), enabling accurate disentangling of albedo in shadowed regions.

\noindent \textbf{Complex Color Loss}
To prevent $\pdi{}(\prir{})$ from converging to 0.5 for all regions, and to enable more confident direct lighting probabilities, we similarly apply a ``complex illumination" loss for $W_{\text{comp}}(\prir{})$ values near 0.5: 

\begin{equation}
\mathcal{L}_{\text{ci}} = \lambda_{\text{ci}} \cdot W_{\text{comp}}(\prir{})
\end{equation}

This loss term helps disambiguate direct lighting probabilities by penalizing values near 0.5; for challenging-geometry regions that cannot be properly resolved, the estimated complex illumination component enables more nuanced shading and soft shadows.

\noindent \textbf{Normal Estimation} We estimate normal vector fields for our scenes following \cite{verbin2022ref,boss2021nerd,srinivasan2021nerv}, using the gradient of the volume density with respect to position: 

\begin{equation}
\hat{\mathbf{n}}(\mathbf{x}) = - \frac{\nabla \sigma(\mathbf{x})}{\lVert \nabla \sigma(\mathbf{x}) \rVert}
\end{equation}

And further predict a 3-vector output $\hat{\mathbf{n}}'_i$ using the output spatial MLP features, applying a regularization term as follows to predict more accurate normal vector fields: 

\begin{equation}
\mathcal{L}_{\text{normal}} = \sum_{i} w_i \lVert \hat{\mathbf{n}}_i - \hat{\mathbf{n}}'_i \rVert^2
\end{equation}

In addition to smoother surface normal predictions, we find this regularization term enables smoother geometry and depth estimation, acting as an effective geometry prior particularly for urban scenes characterized by smooth, cuboidal buildings. 

\noindent \textbf{Transient Loss} 
Satellite imagery captures are both multi-view and multi-time, capturing transient features such as cars, construction events, and seasonal vegetation changes. To account for transient features, we adopt the technique proposed in NeRF-W \cite{martin2021nerf}, predicting $\beta(\mathbf{t}_{\text{i}}, \mathbf{x})$ as a function of per-input transient embedding. We incorporate the NeRF-W loss to ensure $\beta$ uncertainty captures transients without converging to infinity: 

\vspace{-1em} 

\begin{equation}
\mathcal{L}_{\beta} = \sum_{\prir{}} \frac{\lVert \mathbf{c}(\prir{}) - \mathbf{c}_{gt}(\prir{}) \rVert_2^2}{2(\beta(\prir{}) + \beta_{min})^2} + \frac{\log (\beta(\prir{}) + \beta_{min}) + \eta}{2}
\end{equation} 

\begin{table}
\centering
\resizebox{\columnwidth}{!}{%
\begin{tabular}{l|cc|cc|cc|cc}
\toprule
\multicolumn{9}{c}{\textbf{Comparison of Satellite Reconstruction Techniques}} \\ 
\hline 
Approach & \multicolumn{2}{c|}{JAX 004} & \multicolumn{2}{c|}{JAX 068} & \multicolumn{2}{c|}{JAX 214} & \multicolumn{2}{c}{JAX 260}  \\
& PSNR  & SSIM  & PSNR  & SSIM  & PSNR  & SSIM  & PSNR  & SSIM  \\
\hline
S-NeRF   & \cellcolor{tabsecond}21.80 & \cellcolor{tabsecond}0.641 & \cellcolor{tabsecond}21.13 &                      0.691 &  \cellcolor{tabthird}18.70 &                      0.577 &  \cellcolor{tabthird}17.62 &                      0.546 \\
SatNeRF  &  \cellcolor{tabfirst}22.25 &  \cellcolor{tabfirst}0.669 &  \cellcolor{tabthird}20.56 & \cellcolor{tabsecond}0.701 &                      18.55 &  \cellcolor{tabthird}0.600 &                      17.42 &  \cellcolor{tabthird}0.553 \\
EO-NeRF* &                      20.52 &                      0.619 &                      18.37 &  \cellcolor{tabthird}0.693 & \cellcolor{tabsecond}19.92 &  \cellcolor{tabfirst}0.684 & \cellcolor{tabsecond}17.96 &  \cellcolor{tabfirst}0.569 \\
\textbf{\model{}} &  \cellcolor{tabthird}21.24 &  \cellcolor{tabthird}0.634 &  \cellcolor{tabfirst}21.54 &  \cellcolor{tabfirst}0.725 &  \cellcolor{tabfirst}20.06 & \cellcolor{tabsecond}0.679 &  \cellcolor{tabfirst}18.23 & \cellcolor{tabsecond}0.568 \\
\hline
\end{tabular}
}
\caption{\textbf{Evaluation metrics for \model{} compared to previous approaches}. Novel view rendering on real, unseen, \textit{future} satellite images. \model{} provides improvements across test scenes for novel view and lighting synthesis. }
\label{table:eval_metrics}
\end{table}

\begin{table}
\centering
\resizebox{0.8\columnwidth}{!}{%
\begin{tabular}{l|c|c|c|c}
\toprule
\multicolumn{5}{c}{\textbf{Satellite Depth Estimation MAE}} \\ 
\hline 
Approach & \multicolumn{1}{c|}{JAX 004} & \multicolumn{1}{c|}{JAX 068} & \multicolumn{1}{c|}{JAX 214} & \multicolumn{1}{c}{JAX 260}  \\
\hline
S-NeRF   & \cellcolor{tabsecond}1.890 &  \cellcolor{tabthird}1.800 &                      4.510 &                      3.070 \\
SatNeRF  &  \cellcolor{tabfirst}1.508 &  \cellcolor{tabfirst}1.432 &  \cellcolor{tabthird}2.898 &  \cellcolor{tabthird}2.413 \\
EO-NeRF* &                      2.275 &                      1.928 & \cellcolor{tabsecond}2.557 & \cellcolor{tabsecond}1.953 \\
\textbf{\model{}} &  \cellcolor{tabthird}2.194 & \cellcolor{tabsecond}1.738 &  \cellcolor{tabfirst}2.462 &  \cellcolor{tabfirst}1.908 \\
\hline
\end{tabular}
}
\caption{\textbf{Digital surface model estimation for real satellite scenes}. \model{}, trained without initial sun directions, demonstrates comparable depth estimation to previous approaches (that use sun direction), and consistently outperforms EO-NeRF. }
\label{table:depth_eval_metrics}
\end{table}

\section{Experiments} 
\label{sec:eval}

\subsection{Dataset}
We evaluate \model{} on real satellite imagery captured by WorldView-3, a high resolution satellite operating at altitude 617km. The dataset includes images from four locations in Jacksonville, Florida, spanning 2014-2016. Each location has 10-25 multi-date captures at 0.3m resolution. Image metadata was used to verify sun position and capture time. In some cases, multiple image captures of a scene were taken in a single day. To avoid overfitting on one of these instances, we train \model{} on the earliest 75\% of sequentially captured images from each scene, and evaluate on the remaining 25\% of images (captured at subsequent times). Additional details are provided in the supplement. 

\subsection{Implementation Details}

We use bundle-adjusted rational polynomial camera (RPC) models to cast scene sampling rays, following \cite{mari2022sat,gao2021rational}. We train \model{} with a batch size of 2048 rays for 200k iterations, using an initial learning rate of \(5 \times 10^{-4}\) using a step scheduler with decay factor $0.9$. Additional training details are provided in the supplement. 


\subsection{Disentangling Appearance}
We show results for the disentangled outputs of \model{} in Figure \ref{fig:full_dis}. We observe that \model{} can effectively resolve scene albedo, physically-accurate shadows, transient features, scene depth, as well as complex illumination features, which primarily enable shading for challenging geometry regions in trees and water.

\subsection{Comparisons with Baseline}

\begin{figure}[!t]
    \centering
    \includegraphics[width=\columnwidth]{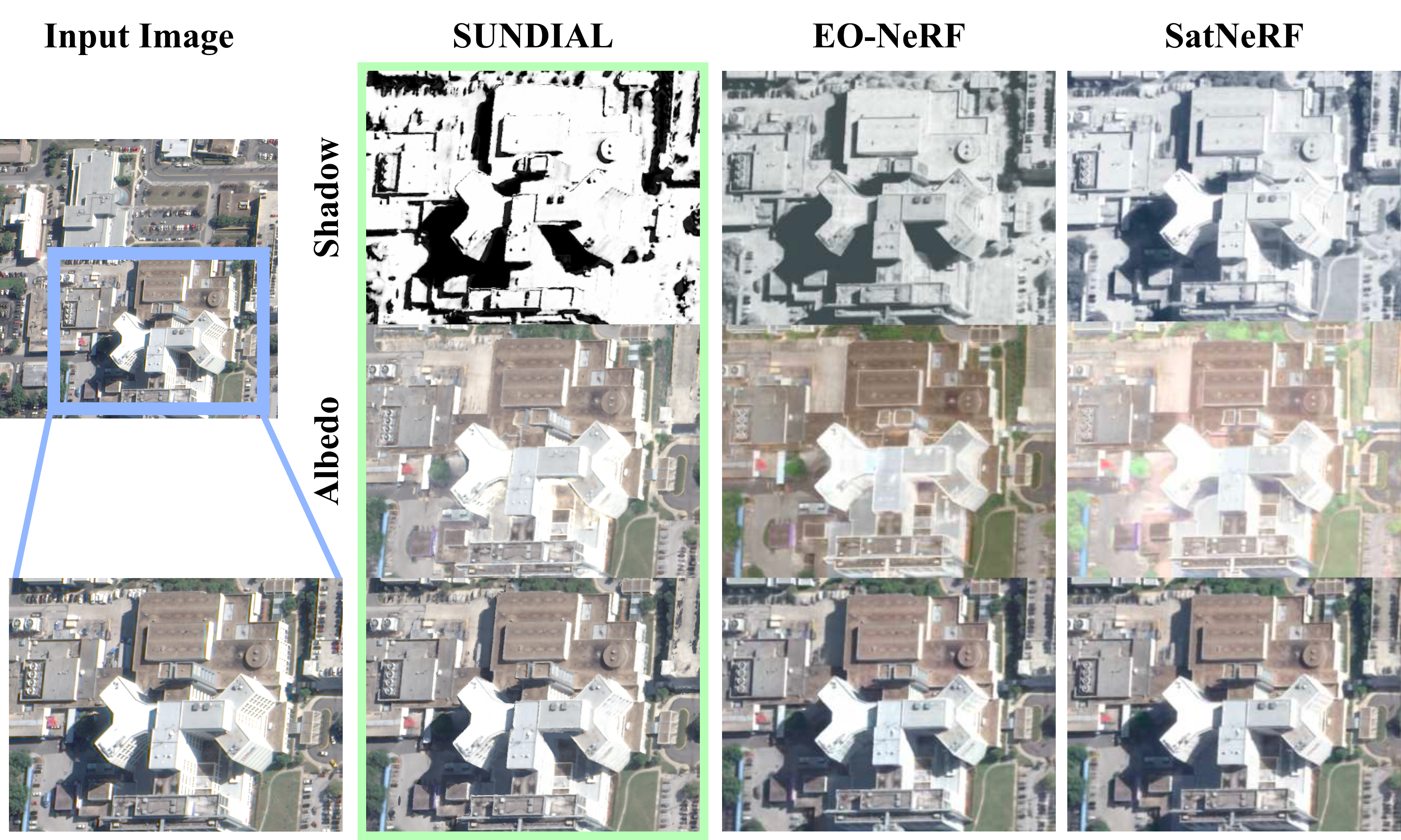}
        \caption{\textbf{Scene disentangling comparison}. \model{} is able to resolve physically-accurate, disambiguated shadows by estimating an additional complex illumination component. EO-NeRF, despite casting geometric shadow rays, cannot properly disentangle soft shadows and shading, instead estimating an ambiguous shading component similar to the non-geometric SatNeRF predictions.}
    \label{fig:can_sat_comp}
\end{figure}

We provide quantitative comparisons to previous approaches in Tables \ref{table:eval_metrics} and \ref{table:depth_eval_metrics}. EO-NeRF does not have a publicly available code release, so we reimplement the proposed model (EO-NeRF*). Our evaluation demonstrates that \model{} accurately disentangles scene geometry, illumination, and sun direction, enabling accurate novel view and lighting synthesis for future, unseen, test images.

We further demonstrate this decomposition accuracy in Figure \ref{fig:can_sat_comp}. \model{} is able to resolve accurate, confident shadow and albedo predictions compared to EO-NeRF and SatNeRF. In particular, SatNeRF shading scalars encompass both shadows and complex illumination components; while EO-NeRF implements a geometric approach to shadow rendering, this simple direct and indirect lighting separation cannot account for these nuanced, soft shadows, resulting in ambiguous shadow predictions.

\begin{figure}[ht]
    \centering
    \begin{subfigure}{.6\columnwidth}
        \includegraphics[width=\linewidth]{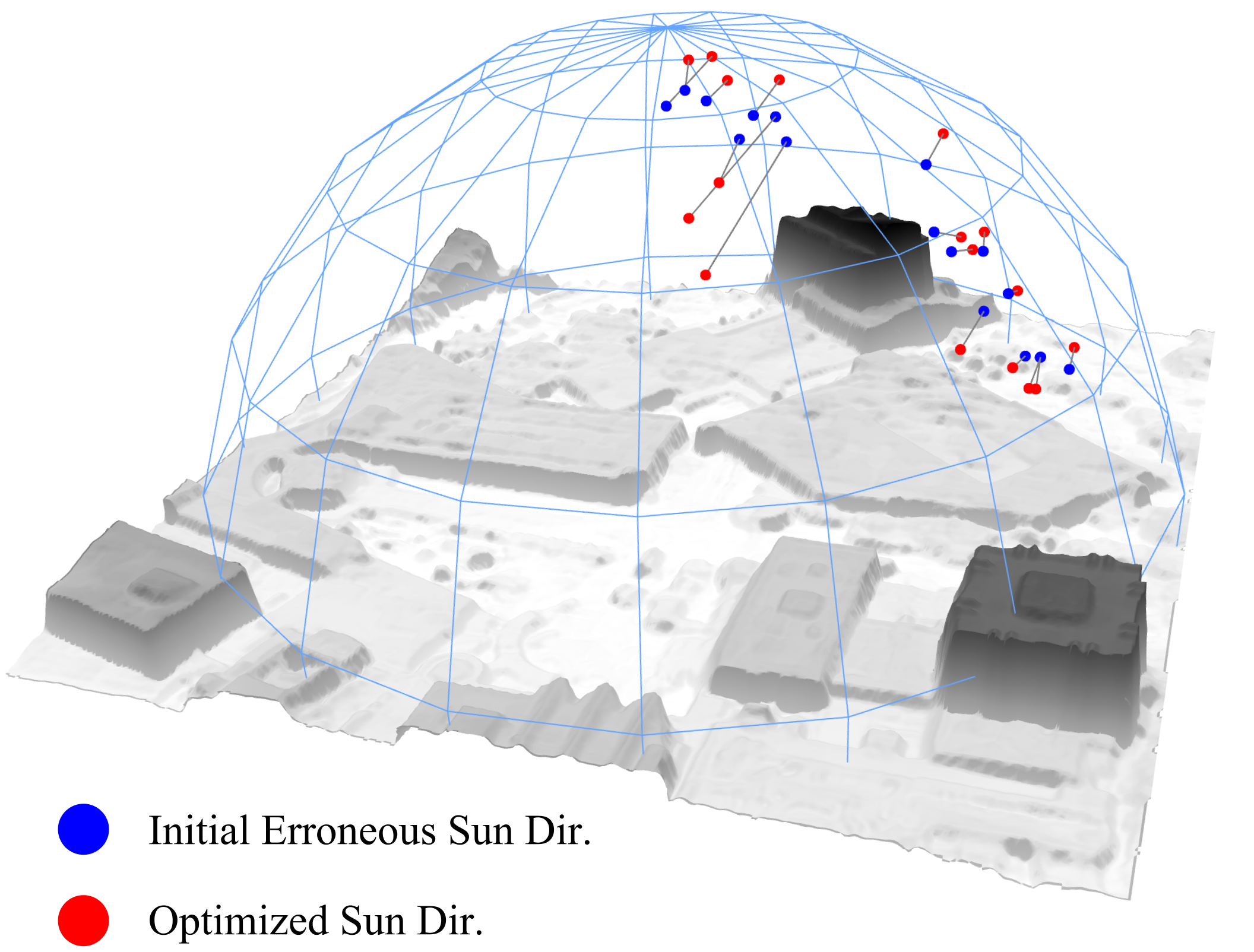}
        \caption{JAX 214 estimated geometry and sun dir.}
    \end{subfigure}
    \hfill
    \begin{subfigure}{.30\columnwidth}
        \centering
        \renewcommand\tabularxcolumn[1]{m{#1}}
        \renewcommand\arraystretch{1.1}
        \setlength\tabcolsep{3pt}
        \resizebox{0.95\columnwidth}{!}{%
            \small
            \begin{tabular}{l c}
                \toprule
                \textbf{Data ID} & \textbf{Sun Dir.} \\
                 & MAE ↓ (°) \\
                \hline
                JAX 004 & \(8.404\) \\
                & \( \scriptsize \pm 3.026\) \\
                \hline
                JAX 068 & \(1.928\) \\
                & \(\pm 1.332\) \\
                \hline
                JAX 214 & \(1.608\) \\
                & \(\pm 1.177\) \\
                \hline
                JAX 260 & \(5.468\) \\
                & \(\pm 4.043\) \\
                \hline
            \end{tabular}
        }
        \caption{Sun dir. MAE}
    \end{subfigure}
    \caption{\textbf{\model{}-estimated scene geometry and sun directions}. \model{} is able to accurately estimate scene geometry jointly with refined sun positions (a); \model{}-estimated sun directions are accurate compared to initial dataset-provided directions (b). }
    \label{fig:geom_and_sun}
\end{figure}



\subsection{Applications}

\noindent\textbf{Sun Estimation} \model{} accurately estimates sun directions jointly with scene geometry (Figure \ref{fig:geom_and_sun}) during training, with reasonable accuracy compared to dataset-provided sun directions. This joint estimation helps disentangle scene geometry and illumination even when sun direction estimations are not available. 

\begin{table}
\centering
\resizebox{0.95\columnwidth}{!}{%
\begin{tabular}{l|ccc|c}
\toprule
\multicolumn{5}{c}{ \textbf{Sun-Based Image Location and Time Estimation} }\\
\hline
Dataset & \multicolumn{2}{c}{Location (lon,lat)} & Error (miles) & Time Error (mins) \\
        & True & Predicted & Great-Circle & L1 Error \\ 
\hline
JAX 004 & (-81.706, 30.358) & (-82.200, 29.611) & 34.86 & \(12.89 \pm 14.34\) \\
JAX 068 & (-81.664, 30.349) & (-81.700, 30.159) & 3.14  & \(6.55 \pm 5.88\) \\
JAX 214 & (-81.664, 30.316) & (-81.576, 30.129) & 6.32  & \(4.97 \pm 4.57\) \\
JAX 260 & (-81.663, 30.312) & (-81.433, 29.820) & 16.67 & \(14.66 \pm 18.35\) \\
\hline
\end{tabular}
}
\caption{\textbf{Image time and location prediction using estimated sun position.} \model{} accurately predicts the time-of-capture and location of images (within a few minutes and miles of ground truth values) using model-estimated sun directions.}
\label{tab:sunest_combined_prediction}
\end{table}

\noindent \textbf{Image Time and Location Estimation}
Given known scene location information, we can use \model{}-estimated sun directions to predict image time-of-capture information. Similarly, given time-of-capture information and \model{}-estimated sun directions, we can estimate image location information. We demonstrate these results in Table \ref{tab:sunest_combined_prediction}. \model{} can estimate image time to within 5 minutes, and location within 4 miles, of ground truth values.

\section{Conclusion}
\label{sec:conclusion}

We present \model{}, a technique for reconstructing 3D satellite-captured scenes by decomposing direct, ambient, and complex illumination effects. Our method can accurately disentangle scene albedo, shadows, ambient illumination, complex illumination, and transient features in satellite imagery. We achieve this by jointly estimating sun direction with scene geometry during training, and casting geometry-based shadow rays that 1) refine hidden scene geometry using oblique sun angles, 2) render separated, geometric shadows, and 3) determine illumination component weights. By disentangling scene geometry and refined illumination components, we enable accurate novel view and lighting synthesis, unlocking new applications in wide-scale environmental modeling. For example, \model{} can help build detailed illumination and transient-aware city models using drone imagery, or render novel scenes with weather features such as fogs, clouds, and weather events that may otherwise inhibit satellite captures. Our approach also demonstrates the importance of drawing insights from remote sensing literature, and can enable future research adopting physics and satellite imagery-based cues to extract hidden information from global satellite captures. 



{\small
\bibliographystyle{ieee_fullname}
\bibliography{egbib}

\begin{thebibliography}{10}\itemsep=-1pt

\bibitem{adamkiewicz2022vision}
Michal Adamkiewicz, Timothy Chen, Adam Caccavale, Rachel Gardner, Preston Culbertson, Jeannette Bohg, and Mac Schwager.
\newblock Vision-only robot navigation in a neural radiance world.
\newblock {\em IEEE Robotics and Automation Letters}, 7(2):4606--4613, 2022.

\bibitem{adeline2013shadow}
Karine~RM Adeline, M Chen, X Briottet, SK Pang, and N Paparoditis.
\newblock Shadow detection in very high spatial resolution aerial images: A comparative study.
\newblock {\em ISPRS Journal of Photogrammetry and Remote Sensing}, 80:21--38, 2013.

\bibitem{barron2022mip}
Jonathan~T Barron, Ben Mildenhall, Dor Verbin, Pratul~P Srinivasan, and Peter Hedman.
\newblock Mip-nerf 360: Unbounded anti-aliased neural radiance fields.
\newblock In {\em Proceedings of the IEEE/CVF Conference on Computer Vision and Pattern Recognition}, pages 5470--5479, 2022.

\bibitem{barron2023zip}
Jonathan~T Barron, Ben Mildenhall, Dor Verbin, Pratul~P Srinivasan, and Peter Hedman.
\newblock Zip-nerf: Anti-aliased grid-based neural radiance fields.
\newblock {\em arXiv preprint arXiv:2304.06706}, 2023.

\bibitem{berra2020advances}
EF Berra and MV Peppa.
\newblock Advances and challenges of uav sfm mvs photogrammetry and remote sensing: Short review.
\newblock In {\em 2020 IEEE Latin American GRSS \& ISPRS Remote Sensing Conference (LAGIRS)}, pages 533--538. IEEE, 2020.

\bibitem{bi2020neural}
Sai Bi, Zexiang Xu, Pratul Srinivasan, Ben Mildenhall, Kalyan Sunkavalli, Milos Hasan, Yannick Hold-Geoffroy, David Kriegman, and Ravi Ramamoorthi.
\newblock Neural reflectance fields for appearance acquisition.
\newblock {\em arXiv preprint arXiv:2008.03824}, 2020.

\bibitem{bishop2011topographic}
Michael~P Bishop and Jeffrey~D Colby.
\newblock Topographic normalization of multispectral satellite imagery.
\newblock {\em Journal of Glaciology}, 55:131--146, 2011.

\bibitem{bonafoni2020albedo}
Stefania Bonafoni and Aliihsan Sekertekin.
\newblock Albedo retrieval from sentinel-2 by new narrow-to-broadband conversion coefficients.
\newblock {\em IEEE Geoscience and Remote Sensing Letters}, 17(9):1618--1622, 2020.

\bibitem{boss2021nerd}
Mark Boss, Raphael Braun, Varun Jampani, Jonathan~T Barron, Ce Liu, and Hendrik Lensch.
\newblock Nerd: Neural reflectance decomposition from image collections.
\newblock In {\em Proceedings of the IEEE/CVF International Conference on Computer Vision}, pages 12684--12694, 2021.

\bibitem{boss2021neural}
Mark Boss, Varun Jampani, Raphael Braun, Ce Liu, Jonathan Barron, and Hendrik Lensch.
\newblock Neural-pil: Neural pre-integrated lighting for reflectance decomposition.
\newblock {\em Advances in Neural Information Processing Systems}, 34:10691--10704, 2021.

\bibitem{Cheng_2020}
Gong Cheng, Xingxing Xie, Junwei Han, Lei Guo, and Gui-Song Xia.
\newblock Remote sensing image scene classification meets deep learning: Challenges, methods, benchmarks, and opportunities.
\newblock {\em {IEEE} Journal of Selected Topics in Applied Earth Observations and Remote Sensing}, 13:3735--3756, 2020.

\bibitem{dare2005shadow}
Paul~M Dare.
\newblock Shadow analysis in high-resolution satellite imagery of urban areas.
\newblock {\em Photogrammetric Engineering \& Remote Sensing}, 71(2):169--177, 2005.

\bibitem{derksen2021shadow}
Dawa Derksen and Dario Izzo.
\newblock Shadow neural radiance fields for multi-view satellite photogrammetry.
\newblock In {\em Proceedings of the IEEE/CVF Conference on Computer Vision and Pattern Recognition}, pages 1152--1161, 2021.

\bibitem{facciolo2017automatic}
Gabriele Facciolo, Carlo De~Franchis, and Enric Meinhardt-Llopis.
\newblock Automatic 3d reconstruction from multi-date satellite images.
\newblock In {\em Proceedings of the IEEE Conference on Computer Vision and Pattern Recognition Workshops}, pages 57--66, 2017.

\bibitem{gao2021rational}
Jian Gao, Jin Liu, and Shunping Ji.
\newblock Rational polynomial camera model warping for deep learning based satellite multi-view stereo matching.
\newblock In {\em Proceedings of the IEEE/CVF International Conference on Computer Vision}, pages 6148--6157, 2021.

\bibitem{gomez2022experimental}
Alvaro G{\'o}mez, Gregory Randall, Gabriele Facciolo, and Rafael~Grompone von Gioi.
\newblock An experimental comparison of multi-view stereo approaches on satellite images.
\newblock In {\em Proceedings of the IEEE/CVF Winter Conference on Applications of Computer Vision}, pages 844--853, 2022.

\bibitem{grodecki2003block}
Jacek Grodecki and Gene Dial.
\newblock Block adjustment of high-resolution satellite images described by rational polynomials.
\newblock {\em Photogrammetric Engineering \& Remote Sensing}, 69(1):59--68, 2003.

\bibitem{guo2010removing}
Jianhong Guo, Lu Liang, and Peng Gong.
\newblock Removing shadows from google earth images.
\newblock {\em International Journal of Remote Sensing}, 31(6):1379--1389, 2010.

\bibitem{hofierka2021estimating}
Jaroslav Hofierka and Katar{\'\i}na Ona{\v{c}}illov{\'a}.
\newblock Estimating visible band albedo from aerial orthophotographs in urban areas.
\newblock {\em Remote Sensing}, 14(1):164, 2021.

\bibitem{irvin1989methods}
R~Bruce Irvin and David~M McKeown.
\newblock Methods for exploiting the relationship between buildings and their shadows in aerial imagery.
\newblock {\em IEEE Transactions on Systems, Man, and Cybernetics}, 19(6):1564--1575, 1989.

\bibitem{li20223d}
Yunzhu Li, Shuang Li, Vincent Sitzmann, Pulkit Agrawal, and Antonio Torralba.
\newblock 3d neural scene representations for visuomotor control.
\newblock In {\em Conference on Robot Learning}, pages 112--123. PMLR, 2022.

\bibitem{liasis2016satellite}
Gregoris Liasis and Stavros Stavrou.
\newblock Satellite images analysis for shadow detection and building height estimation.
\newblock {\em ISPRS Journal of Photogrammetry and Remote Sensing}, 119:437--450, 2016.

\bibitem{liu2012object}
Wen Liu and Fumio Yamazaki.
\newblock Object-based shadow extraction and correction of high-resolution optical satellite images.
\newblock {\em IEEE Journal of Selected Topics in Applied Earth Observations and Remote Sensing}, 5(4):1296--1302, 2012.

\bibitem{liu2023nero}
Yuan Liu, Peng Wang, Cheng Lin, Xiaoxiao Long, Jiepeng Wang, Lingjie Liu, Taku Komura, and Wenping Wang.
\newblock Nero: Neural geometry and brdf reconstruction of reflective objects from multiview images.
\newblock {\em arXiv preprint arXiv:2305.17398}, 2023.

\bibitem{longbotham2011spatial}
Nathan Longbotham, Chad Bleiler, Chuck Chaapel, Chris Padwick, William Emery, and Fabio Pacifici.
\newblock Spatial classification of worldview-2 multi-angle sequence.
\newblock In {\em 2011 Joint Urban Remote Sensing Event}, pages 105--108. IEEE, 2011.

\bibitem{mari2019bundle}
Roger Mar{\'\i}, Carlo de Franchis, Enric Meinhardt-Llopis, and Gabriele Facciolo.
\newblock To bundle adjust or not: A comparison of relative geolocation correction strategies for satellite multi-view stereo.
\newblock In {\em Proceedings of the IEEE/CVF International Conference on Computer Vision Workshops}, pages 0--0, 2019.

\bibitem{mari2022sat}
Roger Mar{\'\i}, Gabriele Facciolo, and Thibaud Ehret.
\newblock Sat-nerf: Learning multi-view satellite photogrammetry with transient objects and shadow modeling using rpc cameras.
\newblock In {\em Proceedings of the IEEE/CVF Conference on Computer Vision and Pattern Recognition}, pages 1311--1321, 2022.

\bibitem{mari2023multi}
Roger Mar{\'\i}, Gabriele Facciolo, and Thibaud Ehret.
\newblock Multi-date earth observation nerf: The detail is in the shadows.
\newblock In {\em Proceedings of the IEEE/CVF Conference on Computer Vision and Pattern Recognition}, pages 2034--2044, 2023.

\bibitem{martin2021nerf}
Ricardo Martin-Brualla, Noha Radwan, Mehdi~SM Sajjadi, Jonathan~T Barron, Alexey Dosovitskiy, and Daniel Duckworth.
\newblock Nerf in the wild: Neural radiance fields for unconstrained photo collections.
\newblock In {\em Proceedings of the IEEE/CVF Conference on Computer Vision and Pattern Recognition}, pages 7210--7219, 2021.

\bibitem{mildenhall2021nerf}
Ben Mildenhall, Pratul~P Srinivasan, Matthew Tancik, Jonathan~T Barron, Ravi Ramamoorthi, and Ren Ng.
\newblock Nerf: Representing scenes as neural radiance fields for view synthesis.
\newblock {\em Communications of the ACM}, 65(1):99--106, 2021.

\bibitem{Mohney_2020}
Doug Mohney.
\newblock Terabytes from space: Satellite imaging is filling data centers, Apr 2020.

\bibitem{mostafa2017review}
Yasser Mostafa.
\newblock A review on various shadow detection and compensation techniques in remote sensing images.
\newblock {\em Canadian journal of remote sensing}, 43(6):545--562, 2017.

\bibitem{ozcanli2014automatic}
Ozge~C Ozcanli, Yi Dong, Joseph~L Mundy, Helen Webb, Riad Hammoud, and Tom Victor.
\newblock Automatic geo-location correction of satellite imagery.
\newblock In {\em Proceedings of the IEEE Conference on Computer Vision and Pattern Recognition Workshops}, pages 307--314, 2014.

\bibitem{polidorio2003automatic}
Airton~Marco Polidorio, Franklin~C{\'e}sar Flores, Nilton~Nobuhiro Imai, Antonio~MG Tommaselli, and Clelia Franco.
\newblock Automatic shadow segmentation in aerial color images.
\newblock In {\em 16th brazilian symposium on computer graphics and image processing (SIBGRAPI 2003)}, pages 270--277. IEEE, 2003.

\bibitem{santosa2016evaluation}
Purnama~Budi Santosa.
\newblock Evaluation of satellite image correction methods caused by differential terrain illumination.
\newblock In {\em Forum Geografi}, volume~30, pages 1--13, 2016.

\bibitem{shepherd2003correcting}
JD Shepherd and JR Dymond.
\newblock Correcting satellite imagery for the variance of reflectance and illumination with topography.
\newblock {\em International Journal of Remote Sensing}, 24(17):3503--3514, 2003.

\bibitem{simeonov2022neural}
Anthony Simeonov, Yilun Du, Andrea Tagliasacchi, Joshua~B Tenenbaum, Alberto Rodriguez, Pulkit Agrawal, and Vincent Sitzmann.
\newblock Neural descriptor fields: Se (3)-equivariant object representations for manipulation.
\newblock In {\em 2022 International Conference on Robotics and Automation (ICRA)}, pages 6394--6400. IEEE, 2022.

\bibitem{srinivasan2021nerv}
Pratul~P Srinivasan, Boyang Deng, Xiuming Zhang, Matthew Tancik, Ben Mildenhall, and Jonathan~T Barron.
\newblock Nerv: Neural reflectance and visibility fields for relighting and view synthesis.
\newblock In {\em Proceedings of the IEEE/CVF Conference on Computer Vision and Pattern Recognition}, pages 7495--7504, 2021.

\bibitem{su2016shadow}
Nan Su, Ye Zhang, Shu Tian, Yiming Yan, and Xinyuan Miao.
\newblock Shadow detection and removal for occluded object information recovery in urban high-resolution panchromatic satellite images.
\newblock {\em IEEE Journal of Selected Topics in Applied Earth Observations and Remote Sensing}, 9(6):2568--2582, 2016.

\bibitem{tancik2022block}
Matthew Tancik, Vincent Casser, Xinchen Yan, Sabeek Pradhan, Ben Mildenhall, Pratul~P Srinivasan, Jonathan~T Barron, and Henrik Kretzschmar.
\newblock Block-nerf: Scalable large scene neural view synthesis.
\newblock In {\em Proceedings of the IEEE/CVF Conference on Computer Vision and Pattern Recognition}, pages 8248--8258, 2022.

\bibitem{tiwary2023orca}
Kushagra Tiwary, Akshat Dave, Nikhil Behari, Tzofi Klinghoffer, Ashok Veeraraghavan, and Ramesh Raskar.
\newblock Orca: Glossy objects as radiance-field cameras.
\newblock In {\em Proceedings of the IEEE/CVF Conference on Computer Vision and Pattern Recognition}, pages 20773--20782, 2023.

\bibitem{tiwary2022towards}
Kushagra Tiwary, Tzofi Klinghoffer, and Ramesh Raskar.
\newblock Towards learning neural representations from shadows.
\newblock In {\em European Conference on Computer Vision}, pages 300--316. Springer, 2022.

\bibitem{verbin2022ref}
Dor Verbin, Peter Hedman, Ben Mildenhall, Todd Zickler, Jonathan~T Barron, and Pratul~P Srinivasan.
\newblock Ref-nerf: Structured view-dependent appearance for neural radiance fields.
\newblock In {\em 2022 IEEE/CVF Conference on Computer Vision and Pattern Recognition (CVPR)}, pages 5481--5490. IEEE, 2022.

\bibitem{vora2021nesf}
Suhani Vora, Noha Radwan, Klaus Greff, Henning Meyer, Kyle Genova, Mehdi~SM Sajjadi, Etienne Pot, Andrea Tagliasacchi, and Daniel Duckworth.
\newblock Nesf: Neural semantic fields for generalizable semantic segmentation of 3d scenes.
\newblock {\em arXiv preprint arXiv:2111.13260}, 2021.

\bibitem{wang2019detection}
Tianxing Wang, Jiancheng Shi, Husi Letu, Ya Ma, Xingcai Li, and Yaomin Zheng.
\newblock Detection and removal of clouds and associated shadows in satellite imagery based on simulated radiance fields.
\newblock {\em Journal of Geophysical Research: Atmospheres}, 124(13):7207--7225, 2019.

\bibitem{weng2012remote}
Qihao Weng.
\newblock Remote sensing of impervious surfaces in the urban areas: Requirements, methods, and trends.
\newblock {\em Remote Sensing of Environment}, 117:34--49, 2012.

\bibitem{xiangli2022bungeenerf}
Yuanbo Xiangli, Linning Xu, Xingang Pan, Nanxuan Zhao, Anyi Rao, Christian Theobalt, Bo Dai, and Dahua Lin.
\newblock Bungeenerf: Progressive neural radiance field for extreme multi-scale scene rendering.
\newblock In {\em European conference on computer vision}, pages 106--122. Springer, 2022.

\bibitem{xie2023snerf}
Ziyang Xie, Junge Zhang, Wenye Li, Feihu Zhang, and Li Zhang.
\newblock S-nerf: Neural radiance fields for street views, 2023.

\bibitem{yang2022s}
Wenqi Yang, Guanying Chen, Chaofeng Chen, Zhenfang Chen, and Kwan-Yee~K Wong.
\newblock S3-nerf: Neural reflectance field from shading and shadow under a single viewpoint.
\newblock {\em Advances in Neural Information Processing Systems}, 35:1568--1582, 2022.

\bibitem{yu2019study}
Ke Yu, Yunhao Chen, Dandan Wang, Zixuan Chen, Adu Gong, and Jing Li.
\newblock Study of the seasonal effect of building shadows on urban land surface temperatures based on remote sensing data.
\newblock {\em remote sensing}, 11(5):497, 2019.

\bibitem{zhang2022iron}
Kai Zhang, Fujun Luan, Zhengqi Li, and Noah Snavely.
\newblock Iron: Inverse rendering by optimizing neural sdfs and materials from photometric images.
\newblock In {\em Proceedings of the IEEE/CVF Conference on Computer Vision and Pattern Recognition}, pages 5565--5574, 2022.

\bibitem{zhang2021physg}
Kai Zhang, Fujun Luan, Qianqian Wang, Kavita Bala, and Noah Snavely.
\newblock Physg: Inverse rendering with spherical gaussians for physics-based material editing and relighting.
\newblock In {\em Proceedings of the IEEE/CVF Conference on Computer Vision and Pattern Recognition}, pages 5453--5462, 2021.

\bibitem{zhang2021nerfactor}
Xiuming Zhang, Pratul~P Srinivasan, Boyang Deng, Paul Debevec, William~T Freeman, and Jonathan~T Barron.
\newblock Nerfactor: Neural factorization of shape and reflectance under an unknown illumination.
\newblock {\em ACM Transactions on Graphics (ToG)}, 40(6):1--18, 2021.

\bibitem{zhao2023review}
Li Zhao, Haiyan Wang, Yi Zhu, and Mei Song.
\newblock A review of 3d reconstruction from high-resolution urban satellite images.
\newblock {\em International Journal of Remote Sensing}, 44(2):713--748, 2023.

\bibitem{zhi2021place}
Shuaifeng Zhi, Tristan Laidlow, Stefan Leutenegger, and Andrew~J Davison.
\newblock In-place scene labelling and understanding with implicit scene representation.
\newblock In {\em Proceedings of the IEEE/CVF International Conference on Computer Vision}, pages 15838--15847, 2021.

\bibitem{rs13040699}
Tingting Zhou, Haoyang Fu, Chenglin Sun, and Shenghan Wang.
\newblock Shadow detection and compensation from remote sensing images under complex urban conditions.
\newblock {\em Remote Sensing}, 13(4), 2021.

\bibitem{zhou2021shadow}
Tingting Zhou, Haoyang Fu, Chenglin Sun, and Shenghan Wang.
\newblock Shadow detection and compensation from remote sensing images under complex urban conditions.
\newblock {\em Remote Sensing}, 13(4):699, 2021.

\bibitem{zhu2018automatic}
Xiaolin Zhu and Eileen~H Helmer.
\newblock An automatic method for screening clouds and cloud shadows in optical satellite image time series in cloudy regions.
\newblock {\em Remote sensing of environment}, 214:135--153, 2018.

\bibitem{zhu2022nice}
Zihan Zhu, Songyou Peng, Viktor Larsson, Weiwei Xu, Hujun Bao, Zhaopeng Cui, Martin~R Oswald, and Marc Pollefeys.
\newblock Nice-slam: Neural implicit scalable encoding for slam.
\newblock In {\em Proceedings of the IEEE/CVF Conference on Computer Vision and Pattern Recognition}, pages 12786--12796, 2022.

\end{thebibliography}
}

\end{document}